\documentclass{cta-author}

{}
{}
{}

\usepackage{multirow}
\usepackage{pifont}
\usepackage{hyperref}
\usepackage{graphics}

\begin{document}

\title{A Transformer-based Approach for Arabic Offline Handwritten Text Recognition}

\author{\au{Saleh Momeni$^{1}$}
\au{Bagher BabaAli$^{2}$}
}

\address{\add{1, 2}{School of Mathematics, Statistics, and Computer Science, University of Tehran, Tehran, Iran}
\email{babaali@ut.ac.ir}}

\begin{abstract}
Handwriting recognition is a challenging and critical problem in the fields of pattern recognition and machine learning, with applications spanning a wide range of domains. In this paper, we focus on the specific issue of recognizing offline Arabic handwritten text. Existing approaches typically utilize a combination of convolutional neural networks for image feature extraction and recurrent neural networks for temporal modeling, with connectionist temporal classification used for text generation. However, these methods suffer from a lack of parallelization due to the sequential nature of recurrent neural networks. Furthermore, these models cannot account for linguistic rules, necessitating the use of an external language model in the post-processing stage to boost accuracy. To overcome these issues, we introduce two alternative architectures, namely the Transformer Transducer and the standard sequence-to-sequence Transformer, and compare their performance in terms of accuracy and speed. Our approach can model language dependencies and relies only on the attention mechanism, thereby making it more parallelizable and less complex. We employ pre-trained Transformers for both image understanding and language modeling. Our evaluation on the Arabic KHATT dataset demonstrates that our proposed method outperforms the current state-of-the-art approaches for recognizing offline Arabic handwritten text.
\end{abstract}

\maketitle

\section{Introduction}
Optical Character Recognition (OCR) is a technology that enables automatic recognition of printed or handwritten text in scanned documents, converting it into machine-readable text. OCR is widely used in applications such as document digitization, search and translation of documents, and autonomous vehicles. OCR tasks typically involve two sub-problems: text detection and text recognition. The text detection algorithms aim to identify text samples in input images, while the text recognition module attempts to comprehend the patch contents and transcribe them into natural language tokens.

Despite significant advances in OCR technology for English text in recent years, other scripts such as Arabic remain relatively unexplored. The Arabic script is widely considered to be one of the most challenging scripts for OCR due to its cursive nature, which presents numerous obstacles, including difficulties with segmentation, character overlapping, and context dependency \citep{ref15}. While previous research in Arabic handwriting recognition has largely focused on individual letters or isolated words \citep{ref27, ref39, ref40}, this approach may not be optimal for dense text, where it can be challenging to identify separate characters or words. To address this issue, this study proposes an end-to-end line-based OCR pipeline to extract Arabic handwritten text and determine the optimal Handwritten Text Recognition (HTR) model. The proposed pipeline aims to overcome the challenges associated with Arabic script by processing text at the line level. This approach considers the overall context of the handwritten text, which is particularly important in Arabic, where the shape of a character can vary depending on its position within a word. By processing text at the line level, the proposed pipeline can better handle the complexities of Arabic handwriting and improve recognition accuracy.

Current text recognition methods commonly use a Convolutional Neural Network (CNN) for feature extraction, followed by a Recurrent Neural Network (RNN) for capturing contextual information. However, RNNs are limited in terms of parallel processing during recognition due to their sequential nature. To overcome this limitation, we propose a Transformer-based pipeline that relies solely on the attention mechanism and does not use recurrence, allowing for more efficient and parallel image recognition.

Our model includes a separate text detection module, similar to ViT \citep{ref29}, which is responsible for identifying text instances in text line images. The input image is adjusted to a specific height while maintaining its aspect ratio, then broken into small patches of the same size and passed to the encoder as input. The encoder uses stacked self-attention layers to convert the input sequence into feature representations. The decoder then uses these encoded sequence features to attempt to decode the text content. We explore two decoding architectures: the Transducer decoder \citep{ref20} and the cross-attention decoder \citep{ref21}.

Connectionist Temporal Classification (CTC) \citep{ref31} is a widely used architecture in handwriting recognition systems, which assumes a monotonic alignment between input and output, simplifying the model. However, the output tokens are assumed to be independent in this architecture, which severely limits the ability of language modeling Consequently, CTC decoders are often paired with an external language model. The integration of an external language model has been observed to enhance the accuracy of the model considerably \citep{ref2}, which implies that decoding should incorporate language rules. The Transducer architecture elegantly addresses this limitation of CTC by incorporating a joiner network and considering both visual and language features when predicting the output. Initially, the Transducer architecture was used in conjunction with RNNs and was therefore referred to as a Recurrent Neural Network Transducer (RNN-T) \citep{ref20}. However, this approach is not limited to RNNs. Recent research in text recognition has demonstrated promising results using transformer architecture \citep{ref1, ref2}. Motivated by these findings, we replaced the conventional RNN-T with the Transformer Transducer. This approach aligns with the current trend of substituting recurrent networks with attention modules. To the best of our knowledge, the use of Transformer Transducer architecture for text recognition has not been previously investigated. For our second architecture, we incorporated a standard Transformer decoder \citep{ref19}. The Transformer decoder is the preferred decoder for many sequence-to-sequence (Seq2Seq) tasks, such as machine translation, and is suitable for sequentially processing images while also possessing the power of language modeling.

Deep neural networks are known to require vast amounts of training data to attain a generalized model. In order to tackle this issue, we have developed a synthetic dataset comprising images generated from Arabic texts along with their corresponding labels. Our models are pre-trained on this data prior to fine-tuning on the original dataset. Additionally, we initialize both the encoder and decoder using pre-trained Transformer models to leverage publicly available resources. Our experiments demonstrate that our approach surpasses the current state-of-the-art in the HTR task without the requirement for complex pre/post-processing steps. In summary, our work makes the following contributions:

\begin{enumerate}
    \item We provide a comprehensive comparison of two distinct Transformer-based model design options to determine the best end-to-end architecture for the HTR task.\\
    \item We investigate the performance of the Transformer Transducer in the HTR task. To the best of our knowledge, this architecture has not been previously employed for text recognition.\\
    \item Our proposed model achieves a new state-of-the-art result on the benchmark dataset without the use of complicated pre/post-processing steps.
\end{enumerate}

The remainder of this paper is organized as follows: we review related works in \hyperlink{section2}{section 2} and introduce our proposed architectures in \hyperlink{section3}{section 3}. In \hyperlink{section4}{section 4}, we elaborate on synthetic data generation, implementation details, and the training scheme. \hyperlink{section5}{Section 5} provides an analysis and comparison of the two proposed pipelines, as well as a comparative evaluation of our approach with other existing methods. Finally, we conclude our work in \hyperlink{section6}{section 6}.

\hypertarget{section2}{}
\section{Related Works}
OCR is a challenging problem in which the input image is typically represented as a sequence that must be mapped to corresponding target labels. The challenge lies in the fact that the alignment between the input and output is often unknown. To address this challenge, CTC has emerged as a popular technique that enables training without the need for exact alignment information. Many text recognition frameworks rely on CTC, such as the end-to-end text recognition architecture proposed by Shi et al. \citep{ref12}, which utilizes a CNN for feature extraction and a RNN for learning spatial dependencies. Another approach, introduced by Bluche and Messina \citep{ref23}, employs gated convolution to enable context-sensitive feature extraction. Furthermore, previous studies have demonstrated the effectiveness of multidimensional RNNs in combination with a CTC decoder \citep{ref32, ref33, ref34}.

While the CTC architecture has been successful in OCR, it still has limitations in terms of language modeling, which can be improved by using an external language model as a post-processing step. To overcome this limitation, attention mechanisms have gained popularity in text recognition tasks \citep{ref11, ref13, ref22, ref24}. Chowdhury and Vig \citep{ref11} introduced an RNN encoder-decoder model for HTR that utilized a CNN backbone. Michael et al. \citep{ref22} further explored different attention mechanisms and positional encodings for improving text recognition accuracy.

In recent years, the Transformer architecture \citep{ref19} has gained significant traction in the domain of text recognition systems. Sheng et al. \citep{ref4} were pioneers in utilizing the Transformer architecture for text recognition. They employed a modality-transform block to convert a 2D image to a 1D sequence, followed by a Transformer encoder-decoder for text recognition. The Transformer architecture's ability to model intricate language rules and facilitate greater parallelization has led to its widespread adoption, replacing traditional RNNs in many text recognition systems \citep{ref5, ref25, ref28}. Baek et al. \citep{ref14} proposed a four-stage framework to compare recognizer models and addressed inconsistencies between training and evaluation datasets. Diaz et al. \citep{ref2} conducted research on developing a universal architecture for handwritten and scene text recognition and compared various encoder/decoder combinations. The use of image Transformers \citep{ref10, ref35, ref36} for feature extraction in text recognition tasks has also gained popularity, with Atienza \citep{ref6} and Li et al. \citep{ref1} among the early adopters of ViT-style Transformers, replacing the conventional CNN backbone. While Atienza \citep{ref6} combined the vision Transformer with a CTC decoder, Li et al. \citep{ref1} employed the Seq2Seq paradigm. The cross-attention architecture employed in our study is similar to the latter work, with the exception that we decompose each input image into a 1D sequence of patches, which allows for a more efficient handling of variable-length inputs.

In this paper, we also explore the feasibility of replacing the conventional CTC decoder with a Transducer decoder in OCR models. The use of a Transducer decoder facilitates end-to-end training of the OCR model, thereby obviating the need for an external language model. The RNN-T loss, first proposed by Graves \citep{ref20}, is central to this end-to-end training. Although it has not been widely used in comparison, recent studies in the field of speech recognition have yielded promising results with the use of RNN-T loss, as demonstrated by Zhang et al. \citep{ref3}. Inspired by this observation, we have employed the Transducer architecture in our HTR model.

\hypertarget{section3}{}
\section{Proposed Model Architecture}
In the context of text recognition systems, two fundamental components are typically employed: a visual feature encoder that extracts salient features from textual images, and a decoder that transcribes the final output based on the extracted features. The present study is concerned with the problem of text transcription and, in particular, explores two distinct decoder choices while utilizing a fixed text detection module.

\begin{figure}[h]
\hypertarget{Fig1}{}
\centering{\includegraphics[width=\columnwidth]{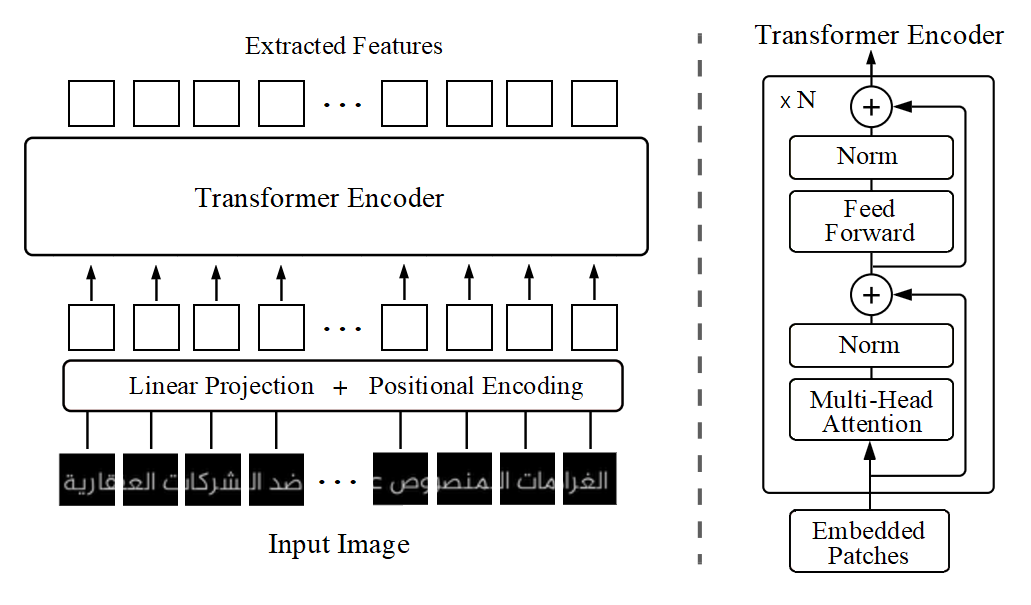}}
\caption{The input image is divided into fixed-size patches. The patches are then embedded, provided with positional encodings, and fed to a standard Transformer encoder.}
\vspace{-17.5pt}
\end{figure}

\subsection{Encoder}
As illustrated in \hyperlink{Fig1}{Fig. 1}, the encoder takes a grayscale image as input and resizes it to a fixed height while maintaining the aspect ratio. Since the transformer architecture cannot process images directly, the input image is partitioned into vertical patches. These patches are then flattened and processed through a linear layer, resulting in a sequence of vectors. To preserve positional information, absolute positional embeddings are added to the vectors, which are learned during the training process. The sequence of vectors is then passed through a sequence of identical transformer layers, each comprising of a self-attention layer and a feed-forward layer. Following each of these layers, a normalizing layer is applied in conjunction with residual connections. The attention mechanism maps a set of queries, keys, and values to the output, which is computed as a weighted average of the values. The weights are calculated based on the similarity between the key and its corresponding query:

\begin{equation}
\text{Attention}(Q,K,V) = \text{softmax}(\frac{QK^T}{\sqrt{d}})V
\end{equation}

Where Q, K, and V are the matrices of the queries, keys, and values respectively, and d is the dimension of the input vector. The self-attention layer employs linear projection to derive the keys, queries, and values from the input sequence. In contrast to a single attention function, each attention module leverages multiple heads with distinct parameters, thereby enabling the model to gather information from distinct representation subspaces:

\begin{equation}
\begin{split}
\text{MultiHead}(Q,K,V) = \text{Concat}(\text{head\textsubscript{1}},\dots, \text{head\textsubscript{h}})W^O \\
\text{where head\textsubscript{i}} = \text{Attention}(QW_i^Q,KW_i^K,VW_i^V)
\end{split}
\end{equation}

The attention mechanism facilitates each encoder layer to encapsulate the information from the entire input sequence in a feature vector, bypassing the necessity for any recurrence.

\subsection{Decoder}
The decoder component of the system takes in the encoded feature sequence and aims to convert the visual cues into corresponding natural language tokens. To achieve this goal, we investigate two distinct decoding methods. The proposed architectures for these decoding methods are illustrated in \hyperlink{Fig2}{Fig. 2} and \hyperlink{Fig3}{Fig. 3}.

\begin{figure}[h]
\hypertarget{Fig2}{}
\includegraphics[width=\columnwidth]{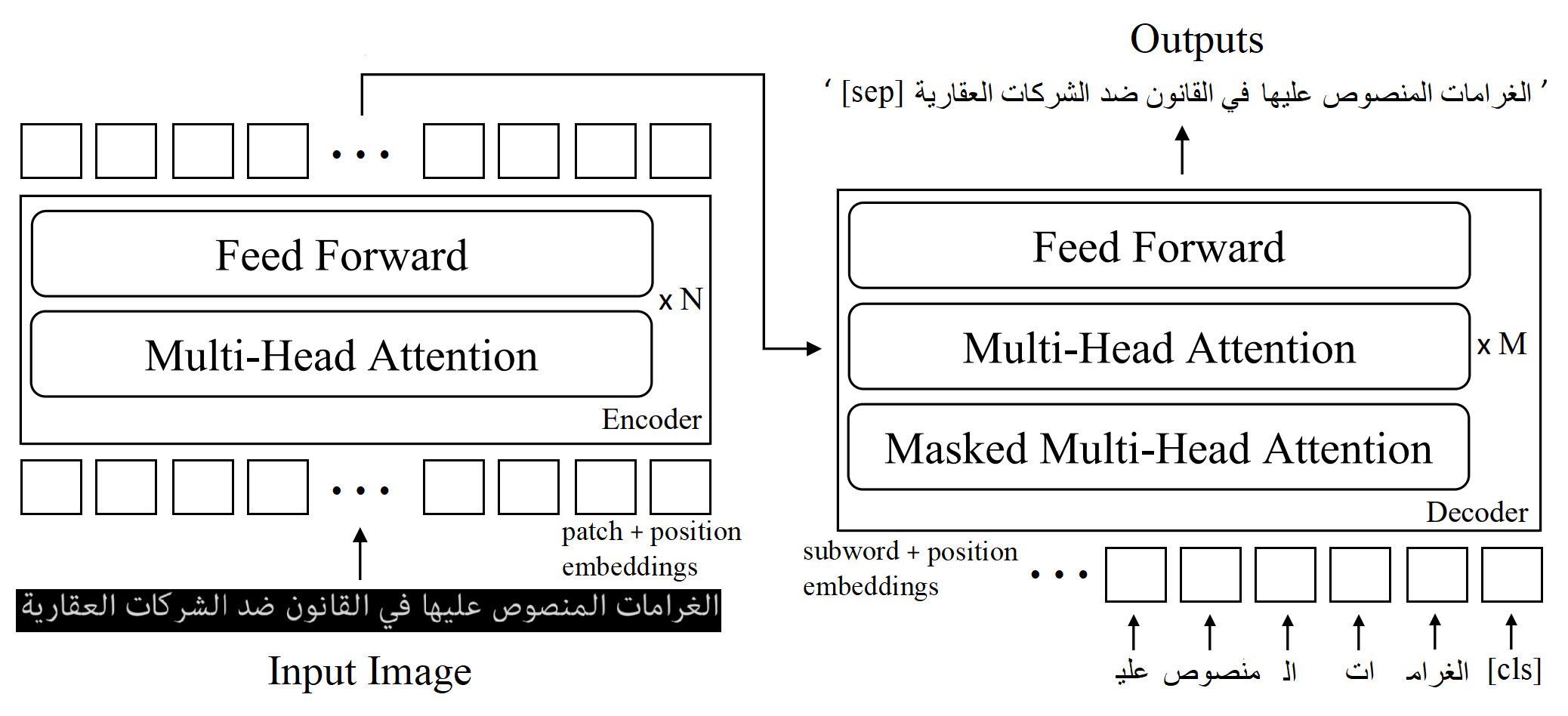}
\centering
\caption{Standard Transformer encoder-decoder architecture, where the decoder attends to the visual features using cross-attention.}
\vspace{-10pt}
\end{figure}

\subsubsection{Transformer with Cross-Attention}
The Transformer decoder takes the encoded feature sequence and generates the output text sequence iteratively, utilizing the previously generated labels as additional input. Each input label is mapped to an embedding vector and positional embeddings are added to these vectors. The resulting embeddings are then passed through several identical transformer decoder layers, each comprising of self-attention and feed-forward layers, as well as a cross-attention layer with separate attention heads to distribute the weights on encoded features. In the cross-attention layer, the queries come from the decoder input, while the keys and values are from the encoder outputs. Attention masking is used during training to ensure that the model does not utilize the information of the tokens that are not yet predicted. To transform the decoder output into a probability distribution over the vocabulary, a linear layer with softmax activation is applied. Finally, the model is trained using cross-entropy loss.

\begin{figure}[h]
\hypertarget{Fig3}{}
\includegraphics[width=\columnwidth]{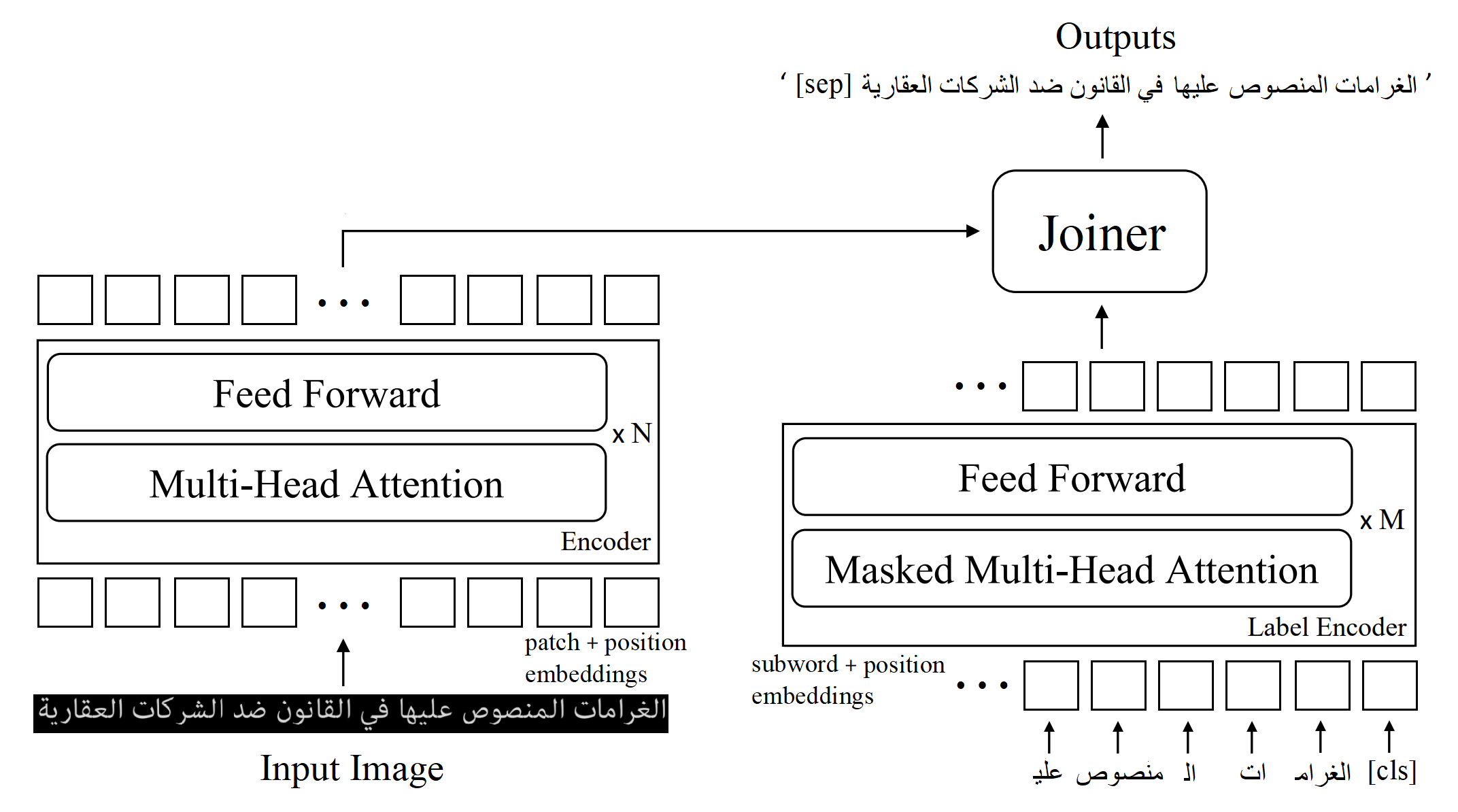}
\centering
\caption{Transformer Transducer architecture, incorporating a joiner network that combines information from the visual encoder and the label encoder to generate the final output.}
\vspace{-5pt}
\end{figure}

\subsubsection{Transformer Transducer}
The Transducer decoder generates a probability distribution over the vocabulary at each time step by receiving the generated feature sequence as input and moving forward on it. This vocabulary incorporates a blank symbol that denotes the absence of output at a given time step. Unlike the CTC, this probability distribution is conditioned on the previously generated tokens. The Transducer decoder is composed of a label encoder and a joiner network. The label encoder resembles a conventional language model that leverages previous outputs to produce features to predict the subsequent label. The label encoder was initially implemented using RNNs in the model proposed by Graves \citep{ref20}. However, in our study, we replaced the RNN with a Transformer architecture. Unlike the cross-attention architecture, the label encoder does not attend to the visual features. Instead, it encodes input labels into a sequence of features that contain semantic information. The last feature vector's information is then used to predict the next label, which is fed to the joiner network along with the visual features. The joiner network adds the predictor vector to the visual features and passes them through a linear layer with softmax activation, generating the probability distribution over the label space. The model can be trained end-to-end using the RNN-T loss. The Transducer model defines the probability of the target sequence \(y\) given the input sequence \(x\):

\begin{equation}
P(y\mid x) = \sum_{\substack{z\in\mathcal{Z}(x, y)}} P(z\mid x)
\end{equation}
\begin{equation}
P(z\mid x) = \prod_i P(z_i\mid x,z_1,\dots,z_{i-1})
\end{equation}

Where \(z\) is an alignment between the input and the target sequence, and \(\mathcal{Z}(x, y)\) is the set of all possible alignments between the two sequences. When utilizing the Transducer decoder, the alignment between the input and target sequences can vary depending on the generated tokens at each time step, leading to multiple possible alignments. As a result, the probability of the target sequence is computed by adding the probabilities of all possible alignments. The model's loss is then calculated as the sum of negative log probabilities over the training examples.

\begin{equation}
\text{loss} = -\sum_i \text{ log }P(y_i\mid x_i)
\end{equation}

 However, directly computing the RNN-T loss by summing the probabilities of all possible alignments would be computationally inefficient. To address this issue, we use the forward-backward algorithm described in \citep{ref20}. This algorithm takes advantage of the overlaps between alignments and stores intermediate computations that can be reused, resulting in improved computational efficiency.

 \subsection{Model Initialization}
 In order to enhance the accuracy of our model, we adopt an approach of initializing both the encoder and decoder with publicly available pre-trained models that have been trained on large-scale datasets. Specifically, we have employed the DeiT \citep{ref10} model as the encoder and the asafaya-BERT \citep{ref9} model as the decoder. The DeiT model is a data-efficient image Transformer that is trained on the ImageNet dataset, and it introduces a distillation token to learn effectively from a teacher, resulting in competitive results on ImageNet without requiring external data. On the other hand, the asafaya-BERT model is the Arabic version of the BERT \citep{ref30} model.

Despite the availability of pre-trained models, we have faced some challenges in initializing our model due to differences in the structure of our models and the pre-trained ones. For instance, the patch embedding layer in our encoder is different from the DeiT model, as the latter is designed to accept RGB images with a resolution of 384x384, while our encoder works with grayscale images of variable length. Moreover, the attention layers in our attention-based decoder that enable the encoder-decoder interactions are not present in the Arabic BERT. Furthermore, our Transducer decoder includes an additional joiner module. To overcome these differences, we have initialized all the parameters mentioned above randomly.

\hypertarget{section4}{}
\section{Experimental Setup}
In this section, we begin by outlining the method employed for generating synthetic data to train the model. We then expound on the benchmark dataset, implementation details, and evaluation criteria used to assess the model's performance.

\subsection{Synthetic Data For Pre-Training}
Deep learning models require a substantial amount of training data to generate a model that can generalize well. To this end, we generated a synthetic dataset comprising 500,000 printed Arabic text-line images, along with their corresponding ground truth, using various open-source fonts. We used text from \citep{ref26} to render the synthetic data. To improve the diversity and realism of the synthetic images, we applied various image processing techniques, such as shearing, rotation, distortion, erosion, and compression. Our models were trained on the synthetic dataset and fine-tuned on the original dataset to evaluate their performance.

\begin{figure}[h]
\hyperlink{Fig4}{}
\includegraphics[width=\columnwidth]{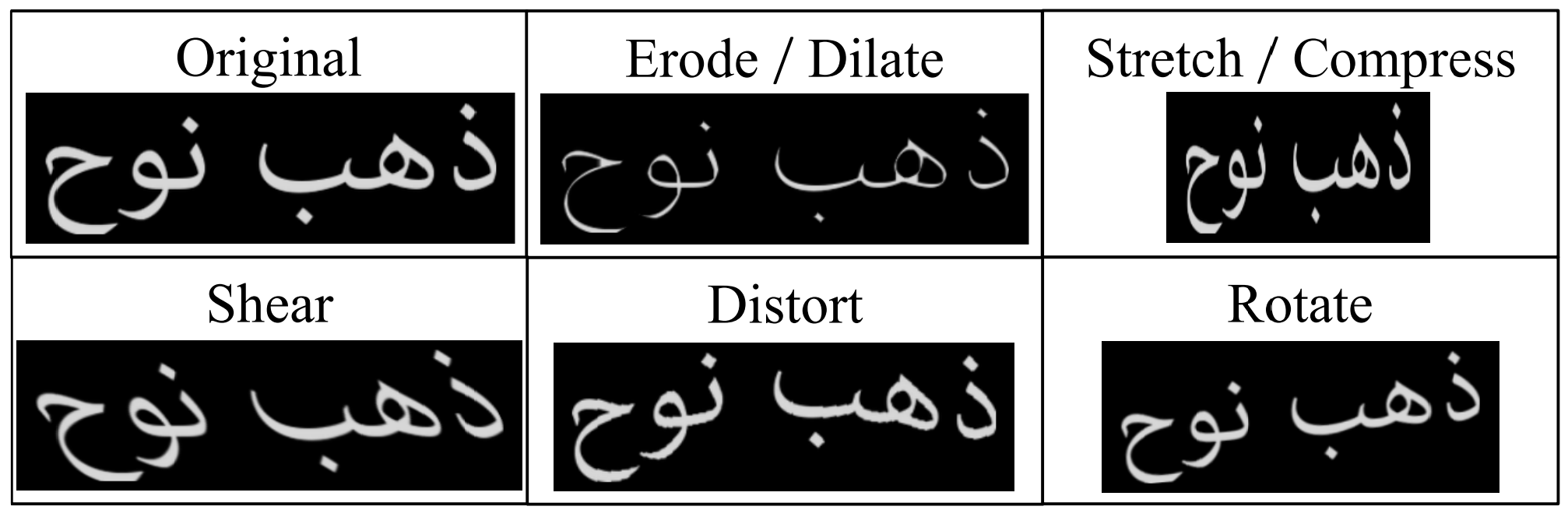}
\centering
\caption{Illustration of image processing techniques used to produce synthetic data.}
\vspace{-17.5pt}
\end{figure}

\subsection{Dataset}
The evaluation of our proposed method is conducted using the KHATT \citep{ref17} dataset, which is publicly available and widely used in the field of Arabic handwriting recognition. The KHATT dataset comprises 1000 handwritten Arabic forms written by various writers from different backgrounds, including age, education levels, and countries. The dataset includes 2000 paragraph images with similar text and 2000 paragraph images with unique text, along with their corresponding ground truth. For the purposes of our experiments, we focus on the unique-text paragraphs, which consist of 6742 text lines after segmentation.

\begin{figure}[h]
\hyperlink{Fig5}{}
\includegraphics[width=\columnwidth]{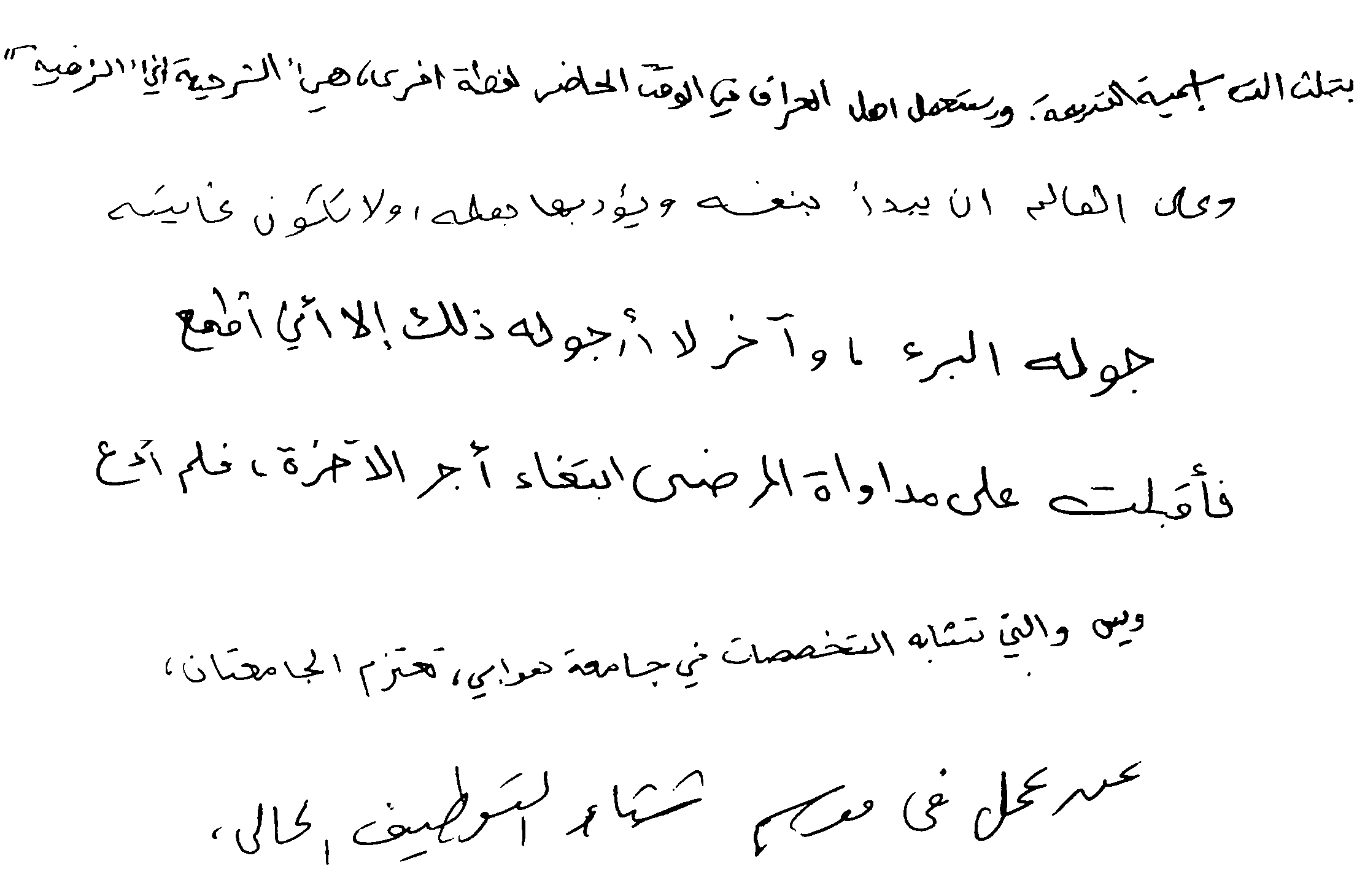}
\centering
\caption{Example text line images in the KHATT dataset which contains handwritten text with various styles.}
\end{figure}

The KHATT dataset contains numerous text-line images that exhibit high skew and extraneous white regions. To prepare these images for height normalization, it is imperative to first remove any excess whitespace. In order to eliminate these additional spaces, we rotate the text-line images, using the same amount of rotation as their skew, but in the opposite direction. This allows us to search the image for black pixels from top to bottom, and discard any extra empty spaces before height normalization.

\subsection{Implementation Details}
In our approach, we utilize gray-scale images for training and evaluation. We normalize the input image to have a height of 64 pixels while preserving the aspect ratio. The encoder's patch size is set to 64x12, and we pad the inputs to a fixed length of 2400 to enable batch processing. We provide a summary of the network configuration in \hyperlink{Table1}{Table 1}. We compare the accuracy by stacking different numbers of encoder and decoder layers. To account for differences in the hidden size of the encoder and decoder, we append a linear layer to the end of the encoder to project feature vectors to the decoder's dimension. We utilize word piece tokenization with a vocabulary size of 32000 to convert text lines to subword tokens.

\begin{table}[h]
\vspace{-4pt}
\hypertarget{Table1}{}
\centering
\begin{minipage}{21em}
\caption{Transformer parameter setup.}
\end{minipage}
\vspace{5pt}
\label{tab:my-table}
\begin{tabular}{lcc}
\hline
\\[-1.85ex]
Parameter                 & Encoder & \multicolumn{1}{l}{Decoder} \\
\\[-1.9ex] \hline \\[-1.9ex]
Hidden Size               & 768     & 512                         \\
Intermediate Size         & 3072    & 2048                        \\
Number of Attention Heads & 12      & 8                           \\
Dropout Ratio             & 0.1     & 0.1                         \\
\\[-1.9ex] \hline
\end{tabular}%
\vspace{-8pt}
\end{table}

The models are implemented in PyTorch \citep{ref7}, and we initialize them using the DeiT\textsubscript{base} and asafaya-BERT\textsubscript{medium} models from the Hugging Face \citep{ref8} repository. All the experiments in the paper are conducted on a Tesla T4 GPU with a 16 GB memory. The Adam optimizer is utilized with a batch size of 16 for training. During the first 5K steps, the learning rate linearly ramps up from 1e-7 to 1e-5, and then remains constant for 20K steps. Following that, it exponentially decays to 1e-6 over the next 5K steps. We fine-tune the models on the KHATT dataset for 10 epochs before evaluating them.

\subsection{Evaluation Metric}
The Character Error Rate (CER) is a widely used evaluation metric for OCR systems. It is defined as the Levenshtein distance between the predicted text and the ground truth, normalized by the number of characters in the ground truth. The Levenshtein distance between two strings is a measure of the difference between them, and it is calculated as the minimum number of character substitutions, deletions, and insertions required to transform one string into the other. More formally, the CER can be calculated as:

\begin{equation}
CER = \frac{s+i+d}{n}
\end{equation}

where \(s\) is the number of substitutions, \(i\) is the number of insertions, \(d\) is the number of deletions, and \(n\) is the total number of characters in the ground truth.

\hypertarget{section5}{}
\section{Results}
\subsection{Ablation Study}
\vspace{-4pt}
In this section, we conduct a thorough ablation study of our models to investigate the impact of each architectural component on the final performance. We report on key metrics, including accuracy, latency, and the number of parameters to provide insight into the trade-offs associated with our approach.

\begin{table}[h]
\vspace{-4pt}
\hypertarget{Table2}{}
\centering
\caption{CER on the KHATT dataset for models trained from scratch with different numbers of encoder and decoder layers.}
\vspace{5pt}
\resizebox{\columnwidth}{!}{%
\begin{tabular}{cccc}
\hline \\[-1.85ex]
Model & Number of Hidden Layers & \multicolumn{1}{l}{Parameter Size} & \multicolumn{1}{l}{CER(\%)} \\
\\[-1.9ex] \hline \\[-1.9ex]
 & 8 encoder / 12 decoder & 129.0 M & 22.92 \\
Transformer-T & 10 encoder / 10 decoder & 136.9 M & 22.20 \\
 & 12 encoder / 8 decoder & 144.7 M & 21.78 \\
 \\[-1.9ex] \hline \\[-1.9ex]
\multirow{3}{2.25cm}{\centering Transformer with Cross-Attention} & 8 encoder / 12 decoder & 141.6 M & 20.71 \\
 & 10 encoder / 10 decoder & 147.4 M & 20.28 \\
 & 12 encoder / 8 decoder & 153.1 M & 19.99 \\
 \\[-1.9ex] \hline
\end{tabular}%
}
\vspace{-6pt}
\end{table}

\noindent\textbf{Number of encoder and decoder layers analysis:}
We begin by investigating the effect of the number of encoder and decoder layers on the model's accuracy. To do so, we compared three different configurations, each with the same total number of layers. All models were trained from scratch, and beam search was utilized to obtain the final predictions. The results are presented in \hyperlink{Table2}{Table 2}. Our analysis indicated that the model with a higher number of encoder layers outperformed the others in terms of accuracy, suggesting that image understanding is more complex than language modeling in the HTR task. Based on these findings, we selected a configuration of 12 encoder layers and 8 decoder layers for our primary models.

\begin{table}[h]
\vspace{-4pt}
\hypertarget{Table3}{}
\centering
\caption{Comparison of Transformer Transducer and cross-attention based Transformer in terms of accuracy and speed on KHATT test set.}
\vspace{5pt}
\resizebox{\columnwidth}{!}{%
\begin{tabular}{ccccc}
\hline \\[-1.85ex]
Model & Parameter Size & \multicolumn{1}{l}{Beam Width} & \multicolumn{1}{l}{Latency (ms)} & \multicolumn{1}{l}{CER(\%)} \\
\\[-1.9ex] \hline \\[-1.9ex]
 &  & K = 1 & 94.41 & 21.15 \\
Transformer-T & 144.7 M & K = 3 & 118.7 & 19.91 \\
 &  & K = 5 & 132.8 & 19.76 \\
 \\[-1.9ex] \hline \\[-1.9ex]
\multirow{3}{2.25cm}{\centering Transformer with Cross-Attention} & \multirow{3}{*}{153.1 M} & K = 1 & 157.4 & 19.74 \\
 &  & K = 3 & 202.9 & 18.63 \\
 &  & K = 5 & 223.7 & 18.45 \\
 \\[-1.9ex] \hline
\end{tabular}%
}
\vspace{-6pt}
\end{table}

\noindent\textbf{Architecture comparison:}
A comparison was conducted between the Transformer Transducer and the cross-attention Transformer architectures, where both models had identical layers, except for the cross-attention layers present in the Transformer decoder, resulting in a substantial increase of 8.4 million parameters in the cross-attention Transformer compared to the Transducer. It is noteworthy that the joiner network in the Transducer is equivalent to the linear layer at the end of the Transformer decoder, as the label encodings are added to the visual features instead of being concatenated. Evaluation was performed on the KHATT dataset, and the results are presented in \hyperlink{Table3}{Table 3}. The Transformer Transducer achieved slightly lower accuracy than the cross-attention Transformer, but significantly outperformed the latter in terms of latency. The reported latencies represent the average time required by each model to recognize a single text line image in the KHATT test set. It was observed that the Transformer encoder took approximately 15.6 milliseconds to produce the visual features, and most of the latency was associated with the autoregressive nature of the decoders. In general, both architectures were competitive, but there were trade-offs. The cross-attention Transformer had a better CER, which was 1.31\% lower than the Transformer Transducer. However, the Transformer Transducer was about 68\% faster than its cross-attention counterpart.

\begin{table}[h]
\vspace{-4pt}
\hypertarget{Table4}{}
\centering
\caption{Impact of different model components on accuracy: in order to determine the significance of individual components, CER was assessed multiple times by toggling these components on and off.}
\vspace{5pt}
\label{tab:my-table}
\resizebox{\columnwidth}{!}{%
\begin{tabular}{ccccc}
\hline \\[-1.85ex]
Model & Initialization & Beam Search & Augmentation & CER(\%) \\
\\[-1.9ex] \hline \\[-1.9ex]
\multirow{4}{*}{Transformer-T} & \ding{55} & \ding{51} & \ding{51} & 21.78 \\
 & \ding{51} & \ding{55} & \ding{51} & 21.15 \\
 & \ding{51} & \ding{51} & \ding{55} & 20.69 \\
 & \ding{51} & \ding{51} & \ding{51} & 19.76 \\
 \\[-1.9ex] \hline \\[-1.9ex]
\multirow{4}{2.25cm}{\centering Transformer with Cross-Attention} & \ding{55} & \ding{51} & \ding{51} & 19.99 \\
 & \ding{51} & \ding{55} & \ding{51} & 19.74 \\
 & \ding{51} & \ding{51} & \ding{55} & 19.30 \\
 & \ding{51} & \ding{51} & \ding{51} & 18.45 \\
 \\[-1.9ex] \hline
\end{tabular}%
}
\vspace{-6pt}
\end{table}

\noindent\textbf{Importance of model initialization:}
Pre-trained vision and language Transformers were employed as the initial point for our models. Pre-trained weight initialization allows the model to adapt quickly to new tasks with better performance. The significance of model initialization was assessed by training the model from scratch. When trained from scratch, a reduction of 1.54\% in the cross-attention Transformer was observed, with a more notable decline of 2.02\% in the Transformer Transducer. The results concluded that utilizing the pre-existing knowledge from pre-trained models significantly improved the models' accuracy.

\vspace{10pt}
\noindent\textbf{Importance of data augmentation:}
Data augmentation is a widely used technique for reducing overfitting in machine learning models. In this work, various data augmentation techniques, such as shearing, rotation, blurring, and noise, were applied during the fine-tuning process. The primary objective of data augmentation was to generate slightly modified copies of the existing data, which could help to improve the generalization capability of the model. The experimental results illustrate that data augmentation plays a critical role in achieving good performance in our models. Specifically, the performance of the Transformer Transducer and cross-attention Transformer models decreased by 0.93\% CER and 0.85\% CER, respectively, when data augmentation was not applied. These findings emphasize the importance of data augmentation in our work and highlight the need for its inclusion in related applications.

\begin{figure}[h]
\vspace{-4pt}
\hypertarget{Fig6}{}
\includegraphics[width=\columnwidth]{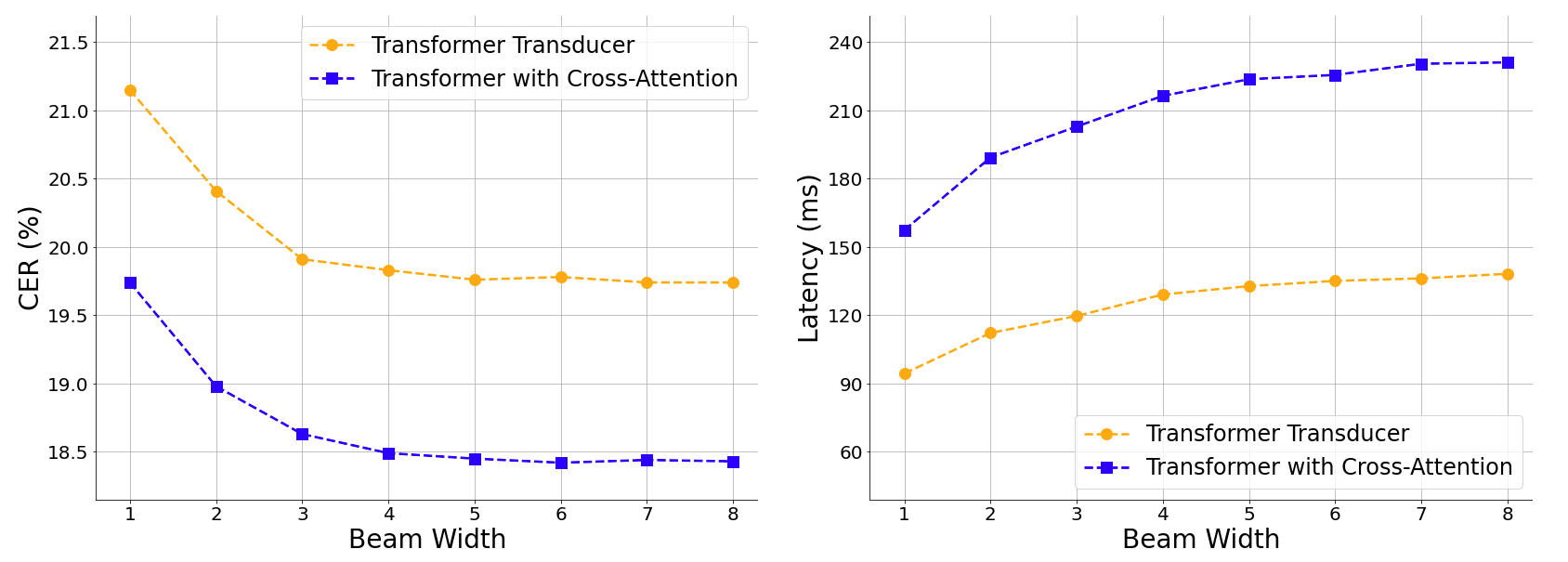}
\centering
\caption{The impact of beam search on accuracy and latency was assessed by gradually increasing the beam width while monitoring changes in CER.}
\vspace{-8pt}
\end{figure}

\noindent\textbf{Influence of beam search:}
we investigated the impact of beam search on the performance of our model. Beam search is an optimized version of the breadth-first search algorithm, which reduces the required memory by using a parameter called beam width. The beam width indicates the number of vertices stored in each step of the algorithm. The effect of beam width on CER and latency was evaluated, and the results were presented in \hyperlink{Fig6}{Fig. 6}. We observed a 1.39\% CER improvement in the Transformer Transducer and a 1.29\% improvement in the cross-attention Transformer by using beam search with a width of K=5, while the decoding time increased by 41\%. Increasing the beam width further did not yield any significant improvement. Our findings revealed that beam search can significantly increase recognition accuracy, albeit at the cost of additional decoding time.

\begin{table}[h]
\vspace{-6pt}
\hypertarget{Table5}{}
\centering
\caption{The results of the evaluations conducted on our models and existing approaches, utilizing the KHATT dataset.}
\vspace{5pt}
\resizebox{\columnwidth}{!}{%
\begin{tabular}{ccccc}
\hline \\[-1.85ex]
\multicolumn{1}{c}{Model} & Architecture & \multicolumn{1}{l}{Recognition Level} & \multicolumn{1}{l}{External LM} & \multicolumn{1}{l}{CER(\%)} \\
\\[-1.9ex] \hline \\[-1.9ex]
Mahmoud et al. \citep{ref38} & HMM & Character & No & 53.87 \\
Ahmad and Fink \citep{ref37} & HMM & Character & No & 41.21 \\
Ahmad et al. \citep{ref16} & MDLSTM + CTC & Character & No & 24.20 \\
\\[-1.9ex] \hline \\[-1.9ex]
\multirow{3}{*}{Ours} & Transformer-T & Subword & No & 19.76 \\ 
\\[-1.9ex]
 & \multirow{2}{2.25cm}{\centering Transformer with Cross-Attention} & \multirow{2}{*}{Subword} & \multirow{2}{*}{No} & \multirow{2}{*}{18.45} \\
 &  &  &  &  \\
\\[-1.9ex] \hline
\end{tabular}%
}
\vspace{-10pt}
\end{table}

\subsection{Comparative Evaluation}
\vspace{-3pt}
The comparison of our proposed models with existing approaches on the KHATT dataset is presented in \hyperlink{Table5}{Table 5}. The results demonstrate that our model outperforms the previous methods and achieves a new state-of-the-art on the KHATT dataset without relying on any complex pre/post processing steps.

\hypertarget{section6}{}
\section{Conclusion}
This study investigated two distinct end-to-end architectures for recognizing Arabic handwritten text lines: the Transformer Transducer and the standard Transformer architecture employing cross-attention. We utilized pre-trained text and image transformers to establish our models. Our proposed method is non-recurrent and open-vocabulary, and both architectures can model complex language dependencies without requiring an external language model. We found that both models are highly competitive, with the cross-attention Transformer achieving superior accuracy and the Transformer Transducer demonstrating faster processing speed. The experimental results indicate that our model achieves a new state-of-the-art performance on the KHATT benchmark.


%


\bibliographystyle{chicago}
\bibliography{Author_tex}

\end{document}